\documentclass[journal]{IEEEtran}

\IEEEoverridecommandlockouts                              

\usepackage{cite}
\usepackage{amsmath,amssymb,amsfonts}
\usepackage{graphicx}
\usepackage{textcomp}
\usepackage{multirow}
\usepackage{cuted}
\usepackage{capt-of}
\usepackage{float}
\usepackage{booktabs}
\usepackage{threeparttable}
\usepackage{mathtools}
\usepackage[mathscr]{eucal}
\usepackage[table,xcdraw]{xcolor}
\usepackage{colortbl}

\usepackage{balance}
\usepackage{color}

\usepackage{marvosym}
\usepackage{indentfirst}
\usepackage{makecell}

\usepackage{algorithm}
\usepackage{algorithmicx}
\usepackage{algpseudocode}

\usepackage{makecell}

\usepackage{subcaption} 

\makeatletter
\let\NAT@parse\undefined
\makeatother
\usepackage{hyperref}

\title{\LARGE \bf
World4RL: Diffusion World Models for Policy Refinement with Reinforcement Learning for Robotic Manipulation
}

\author{Anonymous}

\author{%
Zhennan Jiang$^{1,2,3*}$, Kai Liu$^{1,2*}$, Yuxin Qin$^{1,2}$, Shuai Tian$^{1,2}$, Yupeng Zheng$^{1,2}$ \\
Mingcai Zhou$^{4}$, Chao Yu$^{5,3}$, Haoran Li$^{1,2\dagger}$, Dongbin Zhao$^{1,2,3}$ %
\thanks{$^{1}$ The State Key Laboratory of Multimodal Artificial Intelligence Systems, Institute of Automation, Chinese Academy of Sciences, Beijing, China}%
\thanks{$^{2}$ School of Artificial Intelligence, University of Chinese Academy of Sciences, Beijing, China}%
\thanks{$^{3}$ Zhongguancun Academy, Beijing, China}%
\thanks{$^{4}$ Beijing Zhongke Huiling Robot Technology Co, Beijing, China}%
\thanks{$^{5}$ Department of Electronic Engineering, Tsinghua University, Beijing, China}%
\thanks{$*$ Equal contribution.}%
\thanks{$\dagger$ Corresponding author.}%
}

\begin{document}

\maketitle
\thispagestyle{empty}
\pagestyle{empty}

\begin{abstract}

Robotic manipulation policies are commonly initialized through imitation learning, but their performance is limited by the scarcity and narrow coverage of expert data. Reinforcement learning can refine polices to alleviate this limitation, yet real-robot training is costly and unsafe, while training in simulators suffers from the sim-to-real gap. Recent advances in generative models have demonstrated remarkable capabilities in real-world simulation, with diffusion models in particular excelling at generation. This raises the question of how diffusion model-based world models can be combined to enhance pre-trained policies in robotic manipulation. In this work, we propose World4RL, a framework that employs diffusion-based world models as high-fidelity simulators to refine pre-trained policies entirely in imagined environments for robotic manipulation. Unlike prior works that primarily employ world models for planning, our framework enables direct end-to-end policy optimization. World4RL is designed around two principles: pre-training a diffusion world model that captures diverse dynamics on multi-task datasets and refining policies entirely within a frozen world model to avoid online real-world interactions. We further design a two-hot action encoding scheme tailored for robotic manipulation and adopt diffusion backbones to improve modeling fidelity. Extensive simulation and real-world experiments demonstrate that World4RL provides high-fidelity environment modeling and enables consistent policy refinement, yielding significantly higher success rates compared to imitation learning and other baselines.

\end{abstract}

\section{INTRODUCTION}

Despite recent progress in robotic manipulation, the field still faces challenges for practical deployment. Imitation learning is widely used to bootstrap policies from demonstrations, but its effectiveness is constrained by the inconsistency\cite{XuC1-RSS-25} and limited diversity\cite{SerkanCabi2019ScalingDR,cui2024gapartmanip,liu2025sample} of available datasets. Although offline reinforcement learning (RL) can extract better policies from imperfect data, its susceptibility to overestimation\cite{NEURIPS2019_c2073ffa} still makes it difficult to work effectively with limited datasets. Online RL offers a natural way to refine such pre-trained policies through interaction. However, real-robot RL, while capable of overcoming dataset limitations, suffers from high interaction costs and significant safety risks that hinder large-scale training. Training in simulation avoids these risks but inevitably introduces discrepancies from real-world physics, leading to a persistent sim-to-real gap\cite{DBLP:conf/ssci/ZhaoQW20}.

In recent years, generative models have achieved remarkable progress in the visual domain\cite{DBLP:conf/nips/KarrasALHHLA21}, with diffusion models\cite{Ho2020denoising} demonstrating particularly strong performance in image\cite{chen2025s2guidancestochasticselfguidance} and video generation\cite{HoSGC0F22, chen2025tevir}. Such generative capacity opens new opportunities for modeling complex and dynamic environments, offering a promising path toward learnable world simulators that provide realistic yet flexible environments for RL training in robotic manipulation.

Building on this idea, we introduce World4RL, a framework that systematically integrates diffusion world models into RL for robotic manipulation. World4RL follows a two-stage paradigm: we first pre-train a diffusion world model on multi-task datasets to capture diverse dynamics, and then refine policies entirely within the frozen model to avoid costly and unsafe online interactions. Serving as a high-fidelity simulator, the world model is composed of a diffusion transition model that predicts future observations conditioned on current observations and actions, and a reward classifier that provides sparse success signals, enabling policy optimization without real-world rollouts.

This design of framework contrasts with prior approaches such as IRASim\cite{FangqiIRASim2024} and NWM\cite{bar2024navigationworldmodels}, which primarily use generative video models for planning at test time rather than for direct policy training. A closer line of work, DiWA~\cite{chandra2025diwa}, also employs world models for policy learning. However, it relies on recurrent state-space models (RSSM\cite{hafner2021mastering}), which lead to blurry generations and compounding rollout errors. In contrast, World4RL leverages diffusion backbones that generate sharper and temporally coherent rollouts, thereby supporting effective end-to-end reinforcement learning.

To further adapt world models to robotic manipulation, which involves high-dimensional action spaces and complex environment interactions compared to navigation~\cite{bar2024navigationworldmodels} and games~\cite{alonso2024diffusionworldmodelingvisual}, we investigate two critical design choices: a two-hot action encoding scheme that provides an efficient representation of continuous actions while enabling lossless reconstruction, thereby serving as a robust bridge between the RL agent and the world model, and diffusion backbone architectures that determine the fidelity and consistency of predictions. These considerations are essential for enabling diffusion world models to serve not only as visual predictors but also as reliable simulators for policy training. To this end, our work makes the following key contributions.

\begin{itemize}
    \item We propose World4RL, a two-stage framework for robotic manipulation that pre-trains a diffusion world model and subsequently uses it for RL-based policy refinement.
    \item To improve modeling fidelity and enable more effective policy refinement, we design a two-hot action encoding tailored for robotic manipulation and adopt a diffusion backbone as the world model.
    \item We identify and validate several factors critical to stable policy refinement in imagined environments, including broader action-space coverage and more controlled exploration. Extensive experiments demonstrate that World4RL achieves state-of-the-art results, with 16\% and 25\% absolute gains in simulation and real-robot settings, respectively.
\end{itemize}

\begin{figure*}[t]
    \centering
    \includegraphics[width=\linewidth]{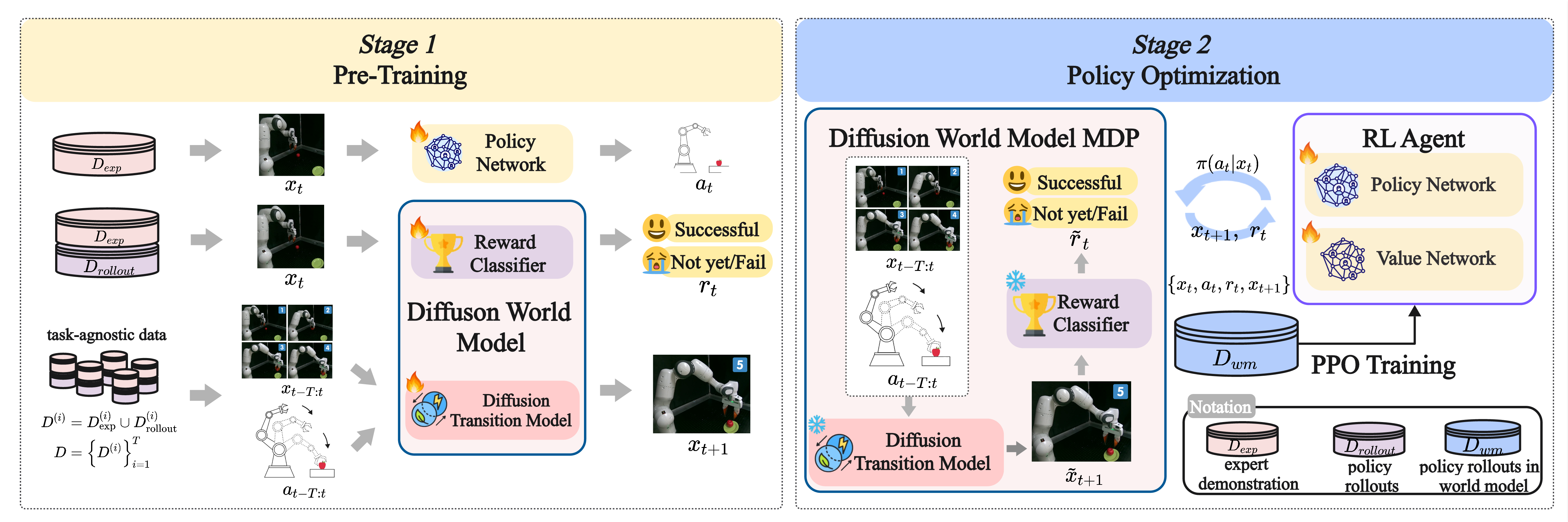}
    \caption{\textbf{Overview of the proposed World4RL framework.} Stage 1 (Pre-training) trains the diffusion transition model on task-agnostic data, optimizes the reward classifier on task-specific success-annotated data, and initializes the policy through imitation learning with expert demonstrations. Stage 2 (Policy Optimization) freezes the pre-trained world model and employs reinforcement learning entirely within imagined rollouts.}
    \captionsetup{skip=3pt}
    \label{fig:framework}
\end{figure*}

\section{Related Work}

\subsection{World Models}
World models have developed rapidly, driven by advances in generative modeling. Early works like VAE-RNN world model~\cite{Ha2018Recurrentworldmodels} inspired latent dynamics models such as DreamerV3~\cite{hafner2024masteringdiversedomainsworld} and TD-MPC2~\cite{hansen2024tdmpc2}, while more recent Transformer-based approaches (e.g., Genie\cite{DBLP:conf/icml/BruceDEPS0LMSAA24}, Drive-WM~\cite{Wang2024DrivingFuture}) significantly extended the temporal horizon and expressiveness. With the rise of conditional diffusion models, world models have achieved high-fidelity video prediction in diverse domains such as autonomous driving (DriveDreamer~\cite{Wang2023DriveDreamer}), navigation (NWM~\cite{bar2024navigationworldmodels}), and gaming (Diamond~\cite{alonso2024diffusionworldmodelingvisual}). These advances indicate that generative world models are becoming powerful tools for simulating complex environments, motivating their application to robotic manipulation.

\subsection{World Models in Robotic Manipulation}

In robotic manipulation, early research primarily adhered to the model-based RL paradigm~\cite{hafner2024masteringdiversedomainsworld, zhao2025seqwm}, where policies must interact with the real environment to generate trajectories for updating the world model. While effective, these methods do not fully exploit the generative capabilities of modern models and still rely on online interaction, limiting their applicability to real-world systems. Recent advancements have shifted toward leveraging world models for planning, where the model predicts trajectories under different action sequences and selects the best one. For example, V-JEPA2~\cite{DBLP:journals/corr/abs-2506-09985} employed self-supervised learning to train a latent action-conditioned world model for robotic planning tasks, while IRASim~\cite{FangqiIRASim2024} introduced a frame-level action-conditioning module into a Diffusion Transformers, significantly improving action responsiveness. DiWA~\cite{chandra2025diwa} employs RSSM~\cite{hafner2021mastering} as its world model to optimize the policy, where the VAE-based latent space limits image generation quality and consequently constrains the overall performance. In contrast, our World4RL framework applies a diffusion world model together with a two-hot action encoding scheme tailored for robotic manipulation, enabling accurate generation and effective policy refinement entirely within imagined rollouts.

\section{Method}
\label{sec:world4rl}
\subsection{Overall Framework}

Given a policy trained with imitation learning from expert demonstrations, our goal is to further improve its performance in robotic manipulation tasks. To this end, we propose the World4RL framework, which addresses the limitations of reinforcement learning (RL), where training in simulators suffers from the inevitable sim-to-real gap, while training directly on real robots incurs expensive and risky interactions. The core idea is to leverage a high-fidelity diffusion world model that enables agents to improve policies entirely in imagined environments. World4RL consists of three key components:

\begin{itemize}
\item Diffusion Transition Model: Serves as a dynamics approximator, predicting future observations conditioned on current observations and actions.
\item Reward Classifier: Considering that robotic manipulation tasks typically involve sparse rewards, we introduce a binary reward classifier to evaluate imagined rollouts generated by the world model.
\item RL-refined Policy: Initialized with the BC policy to provide a stable starting point, and subsequently refined within the world model using the Proximal Policy Optimization (PPO~\cite{DBLP:journals/corr/SchulmanWDRK17}) algorithm.
\end{itemize} 

The overall training pipeline consists of two stages: pre-training and policy optimization. In the pre-training stage, the diffusion transition model is trained on task-agnostic data to generalize across diverse dynamics, the reward classifier is trained on task-specific data annotated with binary success labels, and the policy is trained via imitation learning to provide a stable initialization. In the policy optimization stage, the pre-trained world model is frozen and used as a simulator, while the policy is refined with PPO under sparse rewards through imagined rollouts. This design improves both sample efficiency and safety, while enabling consistent gains over the initial  policy. An overview of the framework is shown in Fig.~\ref{fig:framework}.



\subsection{Pre-training Stage}

\subsubsection{Policy Pre-training}

We first pre-train the policy through imitation learning. This stage provides an imitation-based initialization from expert demonstrations, ensuring that the learned policy $\pi_{\xi}$ can already execute reasonable actions. 

In practice, the policy is parameterized as a stochastic policy that outputs the mean and standard deviation of an action distribution conditioned on the observation. Given an observation $x_t$, the policy defines
\begin{equation}
\pi_{\xi}(a_t \mid x_t) = \mathcal{N}\!\big(\mu_{\xi}(x_t), \Sigma_{\xi}(x_t)\big).
\end{equation}

Formally, given an expert demonstration dataset $D_{exp}$, the policy is trained by maximizing the log-likelihood of expert actions:
\begin{equation}
\mathcal{L}_{BC}(\xi)
=
-\mathbb{E}_{(x_t, a_t)\sim D_{exp}}
\left[
\log \pi_{\xi}(a_t \mid x_t)
\right].
\label{eq:BCJFunc}
\end{equation}

\subsubsection{Reward Classifer}

To provide a reward signal, we introduce a reward classifier that indicates whether the agent has reached a successful state. Given the next observation $x_{t+1}$, the classifier $C_\psi$ outputs the probability that the next observation is a success, defined as $r:= C_\psi(x_{t+1})$.


To make the classifier robust for distinguishing successful states actually reached by the learned policy, we train it not only on expert demonstrations $D_{exp}$, but also on rollout data collected by the pre-trained policy, denoted as $D_{rollout}$. Architecturally, we employ a pre-trained ResNet18\cite{he2016deep} as the visual backbone for feature extraction. Its parameters are optimized by minimizing the binary cross-entropy loss:
\begin{equation}
\label{eq:RCJFunc}
    \mathcal{L}_C(\psi) = - \frac{1}{N} \sum_{i=1}^N \Big[ r_i \log C_\psi(x_{i}) + (1-r_i)\log\big(1-C_\psi(x_{i})\big) \Big].
\end{equation}


\subsubsection{Diffusion Transition Model}



Diffusion models learn to reverse a gradual noising process and generate data through iterative denoising. Following EDM~\cite{karras2022elucidating}, we adopt the preconditioned denoising formulation
\begin{align}
D_{\theta}(x^{\tau}; \tau, c) = c_{\mathrm{skip}}^{\tau} x^{\tau} + c_{\mathrm{out}}^{\tau} F_{\theta}(c_{\mathrm{in}}^{\tau} x^{\tau}; c_{\mathrm{noise}}^{\tau}, c),
\label{eq:ConditionEdmDenosingFuncSimple}
\end{align}
where $F_{\theta}$ is the learnable network, $\tau$ is the diffusion timestep, and $c$ denotes the conditioning information.

Based on this formulation, we model the world dynamics with a diffusion transition model $D_{\theta}$, which predicts the next observation $x^{0}_{t+1}$ from a finite history of observations $x^{0}_{t-T:t}$\footnote{We use $x^0$ to denote the original observation for consistency with diffusion notation.} and actions $a_{t-T:t}$.

To better handle continuous action inputs within the world model, we adopt a two-hot encoding scheme, inspired by DreamerV3~\cite{hafner2024masteringdiversedomainsworld}. Unlike one-hot discretization~\cite{alonso2024diffusionworldmodelingvisual}, latent-space representation~\cite{lee2024behavior}, or token-based approaches~\cite{pertsch2025fast}, two-hot encoding provides a lossless and differentiable representation without introducing reconstruction errors. For each action dimension $a_i \in \mathbb{R}$, given bin values $\mathcal{B} = \{b_1, \dots, b_K\}$, we map $a_i$ to its two nearest bins:
\begin{align}
& \mathbf{t}_i[k] = \frac{b_{k+1} - a_i}{b_{k+1} - b_k}, \quad 
\mathbf{t}_i[k+1] = \frac{a_i - b_k}{b_{k+1} - b_k},
\end{align}
with $\sum_j \mathbf{t}_i[j] = 1$ and $b_k \leq a_i \leq b_{k+1}$, where $\mathbf{t}_i \in \mathbb{R}^K$ denotes the two-hot weight vector for the $i$-th action dimension. This interpolation-based representation preserves continuity while embedding a lightweight discrete structure. In practice, two-hot encoding achieves fine-grained modeling with a moderate number of bins (e.g., $K=21$) and can be optimized end-to-end with policy networks. We denote the encoded action representation as $z$, and replace the original action $a_{t-T:t}$ with $z_{t-T:t}$ as conditional input to the diffusion model.

In addition, the diffusion transition model is trained on a mixture of datasets to ensure robust dynamics modeling. Specifically, the training data consists of three sources: expert demonstrations $D_{exp}$, rollout trajectories collected by the pre-trained policy $D_{rollout}$, and random rollouts $D_{rand}$ generated by sampling actions uniformly from the action space. The expert demonstrations provide high-quality trajectories that capture successful task execution, while policy rollouts introduce states encountered by the learned policy during execution, which better reflect the distribution of states visited during reinforcement learning. Furthermore, random rollouts expose the model to a broader range of state-action transitions, preventing the world model from overfitting to the limited offline dataset and improving its ability to generalize during imagined rollouts in reinforcement learning.

Taking above into consideration, the diffusion transition model $D_{\theta}$ is formally defined as a denoising process conditioned on historical observations $x^{0}_{t-T:t}$ and encoded actions $z_{t-T:t}$, and  (\ref{eq:ConditionEdmDenosingFuncSimple}) can be rewrote as follows:
\begin{align} 
D_{\theta}(x^{\tau}, x^{0}_{t-T:t}, z_{t-T:t}) = 
& c_{\mathrm{skip}}^{\tau} x^{\tau} \nonumber + \\
& c_{\mathrm{out}}^{\tau} F_{\theta}(c_{\mathrm{in}}^{\tau} x^{\tau}; c_{\mathrm{noise}}^{\tau}, x^{0}_{t-T:t}, z_{t-T:t}). \label{eq:DFunc}
\end{align}

After that, we obtain the training objective:
\begin{align}
\mathcal{L}_{D}(\theta) = 
& \; \mathbb{E}_{{x^{\tau} \sim p_{\tau}}} \Bigg[
   \bigg\|
   \mathbf{F}_\theta\!\left(
   c_{\mathrm{in}}^\tau \mathbf{x}_{t+1}^\tau,\,
   \tau,\, x^{0}_{t-T:t},\, z_{t-T:t}\right) \nonumber \\
& \quad - \frac{1}{c_{\mathrm{out}}^\tau}
   \Big(\mathbf{x}_{t+1}^0 - c_{\mathrm{skip}}^\tau \mathbf{x}_{t+1}^\tau\Big)
   \bigg\|^2
   \Bigg].
\label{eq:LFunc}
\end{align}

In practice, we follow the design principles of EDM~\cite{karras2022elucidating} in selecting the noise schedule and hyperparameters (e.g., the design of $c_{\text{in}}$ and $c_{\text{out}}$). For the network architecture, we employ U-Net 2D to model $F_{\theta}$. 

\begin{algorithm}[ht]
\caption{Algorithm of World4RL (Pre-training and Policy Optimization)}
\label{algo:world4rl}
\textbf{Input:}

$D_{\theta},\, C_{\psi},\, \pi_{\xi},\, V_{\phi}$: diffusion transition model, reward classifier, policy, value network

$\mathcal{D}_{\mathrm{exp}},\, \mathcal{D}_{\mathrm{rollout}},\, \mathcal{D}_{\mathrm{wm}}$: expert demos, pre-trained/random policy rollouts, world-model rollouts buffers

\textbf{Pre-training Stage}
\begin{algorithmic}[1]

\State \textbf{(Policy pre-training)}
\State Sample $(x^0_{t},\, a_{t}) \sim \mathcal{D}_{\mathrm{exp}}$
\State Compute $\mathcal{L}_{BC}(\xi)$ using Eq.~\eqref{eq:BCJFunc}
\State $\xi \leftarrow \xi - \alpha \nabla_{\xi} \mathcal{L}_{BC}(\xi)$ \Comment{update policy}

\State \textbf{(Diffusion transition model pre-training)}
\State Sample $(x^0_{t-T:t},\, a_{t-T:t},\, x^0_{t+1}) \sim \mathcal{D}_{\mathrm{exp}} \cup \mathcal{D}_{\mathrm{rollout}}$
\State $z_{t-T:t} \leftarrow \mathrm{TwoHot}(a_{t-T:t})$
\State Sample $\tau \sim \mathcal{U}[0,\mathcal{T}]$ and construct $x^\tau_{t+1}$ by forward noising of $x^0_{t+1}$
\State Compute $\mathcal{L}_{D}(\theta)$ using Eq.~\eqref{eq:DFunc} and \eqref{eq:LFunc}
\State $\theta \leftarrow \theta - \alpha \nabla_{\theta} \mathcal{L}_{D}(\theta)$ \Comment{update diffusion}

\State \textbf{(Reward classifier pre-training)}
\State Sample $(x^0_{i},\, y_i) \sim \mathcal{D}_{\mathrm{exp}}$ \Comment{$y_i \in \{0,1\}$ success label}
\State Compute $\mathcal{L}_{C}(\psi)$ using Eq.~\eqref{eq:RCJFunc})
\State $\psi \leftarrow \psi - \alpha \nabla_{\psi} \mathcal{L}_{R}(\psi)$ \Comment{update classifier}

\end{algorithmic}

\textbf{Policy Optimization Stage}
\begin{algorithmic}[1]
\For{$\mathrm{epoch}=1 \dots \mathrm{max\_epochs}$}
    \State Observe context $x_{t-T:t}$
    \State $a_t \sim \pi_{\xi}(\cdot \mid x_{t})$; \quad $z_t \leftarrow \mathrm{TwoHot}(a_t)$
    \State $\tilde{x}_{t+1} \leftarrow \text{Sample from } D_{\theta}(\cdot;\, x_{t-T:t},\, z_{t-T:t})$
    \State $r_t \leftarrow C_{\psi}(\tilde{x}_{t+1}) \in \{0,1\}$
    \State $\mathcal{D}_{\mathrm{wm}} \leftarrow \mathcal{D}_{\mathrm{wm}} \cup \big(x_{t},\, a_t,\, r_t,\, \tilde{x}_{t+1}\big)$
    \If{$|\mathcal{D}_{\mathrm{wm}}| \ge \mathrm{ppo\_batch\_size}$}
        \State \Call{UpdatePPO}{$\mathcal{D}_{\mathrm{wm}}$}
        \State $\mathcal{D}_{\mathrm{wm}} \leftarrow \varnothing$
    \EndIf
\EndFor
\end{algorithmic}
\end{algorithm}

\subsection{Policy Optimization Stage}

Our objective is to learn an agent $\pi_{\xi}$ that maximizes the expected cumulative rewards in the world model. To achieve this, we adopt PPO\cite{DBLP:journals/corr/SchulmanWDRK17}, which separately optimizes a policy model $\pi_{\xi}$ and a value model $V_{\phi}$. The policy objective is defined as
\begin{align}
\mathcal{L}_{\text{P}}(\xi) = \mathbb{E}_{t} [ \min ( &  \rho_t(\xi) A_t(x_t, a_{t}), \, \\ \nonumber & \text{clip}(\rho_t(\xi), 1 - \epsilon, 1 + \epsilon) A_t(x_t, a_t) ) ], \label{eq:PPOJFunc}
\end{align}
where $\rho_t(\xi) = \frac{\pi_{\xi}(a_t | x_{t})}{\pi_{\xi_{\text{old}}}(a_t | x_{t})}$ is the probability ratio between the updated policy $\pi_{\xi}$ and the reference (old) policy $\pi_{\xi_{\text{old}}}$ that was used to collect trajectories, $A_t(x_t, a_t)$ is the advantage function, and $\epsilon$ is a hyperparameter controlling the clipping range.  

The value function is optimized with the objective:
\begin{equation}
\mathcal{L}_V(\phi) = \mathbb{E}_t \left[ \left( V_{\phi}(x_{t}) - (r_t + \gamma V_{\phi}(x_{t+1})) \right)^2 \right]. \label{eq:VJFunc}
\end{equation}

During optimization, the frozen diffusion world model $D_{\theta}$ is used to generate imagined rollouts conditioned on past observations $x^0_{t-T:t}$ and encoded actions $z_{t-T:t}$. These synthetic trajectories serve as a substitute for real-world interaction, greatly reducing training costs and avoiding hardware risks. The reward classifier $C_{\psi}$ evaluates each imagined next state, providing a sparse binary reward signal $R(s_t,a_t) \in \{0,1\}$.  

However, policy optimization in the learned world model inevitably encounters out-of-distribution (OOD) actions, ledading unstable optimization. To mitigate this issue, we adopt a controlled policy exploration mechanism, implemented by regularizing the policy exploration during PPO optimization. Concretely, for the policy $\pi_{\xi}(a_t \mid x_t) = \mathcal{N}\!\big(\mu_{\xi}(x_t), \Sigma_{\xi}(x_t)\big)$, we clip the predicted standard deviation $\sigma$ in $\Sigma_{\xi}$ to a conservative upper bound. While a common PPO implementation allows $\sigma \le e^{2}$, we instead impose a much tighter constraint $\sigma \le e^{0}$. This constrained exploration keeps sampled actions closer to the support of the world-model training distribution, thereby reducing out-of-distribution rollouts and improving the reliability of imagined transitions.

The policy $\pi_{\xi}$ interacts with the world model in the following loop:    

\begin{itemize}
\item Given the current observation $x^0_t$, an action $a_t$ is sampled from the policy distribution $\pi_{\xi}(\cdot \mid x^0_t)$ and then discretized via two-hot encoding into $z_t$.
\item The diffusion transition model predicts the next observation $x^0_{t+1}$ conditioned on $x^0_{t-T:t}$ and $z_{t-T:t}$. 
\item The reward classifier evaluates $x^0_{t+1}$ to provide a binary success/failure reward.
\item The policy and value models are updated via PPO\cite{DBLP:journals/corr/SchulmanWDRK17} using the generated rollouts.
\end{itemize}
 
This integration of PPO\cite{DBLP:journals/corr/SchulmanWDRK17} with imagined rollouts allows the policy to efficiently explore and improve in sparse reward settings, while the pre-training ensure stable initialization. Together, these design choices enable World4RL to achieve both sample efficiency and robust performance in robotic manipulation tasks.

\section{Experiments}
\label{sec:experiments}

We conduct extensive experiments to evaluate the effectiveness of World4RL. Our goal is to answer three key questions: 1) Can World4RL accurately model fine-grained robotic manipulation tasks and capture task-specific dynamics? 2) Does World4RL facilitate reinforcement learning by enabling more efficient policy training and achieving superior performance compared with strong baselines, including imitation learning, offline reinforcement learning, and world-model based methods? 3) Can World4RL maintain strong performance when deployed on real robot platforms? 

To assess the capability of World4RL as a generative world model, we adopt three widely used metrics: 
\begin{itemize}
    \item \textbf{LPIPS}\cite{zhang2018unreasonable}: perceptual similarity between predictions and ground truth;
    \item \textbf{FID}\cite{heusel2017gans}: distributional quality of generated images;
    \item \textbf{FVD}\cite{unterthiner2018towards}: video-level consistency capturing both spatial and temporal fidelity.
\end{itemize}

For this part, we compare against three representative video prediction models: \textbf{NWM}\cite{bar2024navigationworldmodels}, a DiT-based dynamics world model designed for temporal sequence modeling, \textbf{iVideoGPT}\cite{wu2024ivideogpt}, an autoregressive transformer framework with compressive tokenization for multimodal prediction, and \textbf{DiWA}\cite{chandra2025diwa}, a RSSM-based world model.


To evaluate the effectiveness of World4RL in policy learning, we adopt task success rate (\textbf{SR}) as the primary metric. We conduct experiments on the Meta-World benchmark~\cite{YuQHJHFL19}. We compare World4RL against a broad spectrum of baselines, all evaluated under the fixed-dataset setting without additional online interaction, including
\begin{itemize}
    \item Imitation learning: behavior cloning (\textbf{BC}) and Diffusion Policy (\textbf{DP})~\cite{ChiFDXCBS23};
    \item Offline RL: \textbf{TD3+BC}~\cite{fujimoto2021minimalist} and \textbf{IQL}~\cite{DBLP:conf/iclr/KostrikovNL22};
    \item World-model-based methods: \textbf{IRASim}~\cite{FangqiIRASim2024}, \textbf{DiWA}~\cite{chandra2025diwa}, and \textbf{TD-MPC2}~\cite{hansen2024tdmpc2}.
\end{itemize}



In addition, we compare against hybrid offline-to-online approaches such as \textbf{Uni-O4}\cite{DBLP:conf/iclr/LeiHLH0X24} and \textbf{RLPD}\cite{DBLP:conf/icml/BallSKL23}, and observe the number of
online samples they require to reach the same level of performance. This directly assesses the sample efficiency of World4RL. 

With the evaluation metrics and baselines established, we now present three sets of experiments to assess the capability of World4RL: modeling robotic dynamics, enhancing policy learning, and transferring to real-world robots.

\subsection{Can World4RL Accurately Simulate Robotic Manipulation Environments?}
\label{sec:exp1}

We collect training data from six representative environments in the Meta-World benchmark. For each task, we gather 50 expert trajectories, 150 trajectories generated by a pre-trained policy, and 30 trajectories from a random policy rollout. Each trajectory contains 50 timesteps. During training, the model is conditioned on a history of four consecutive frames along with their corresponding actions, and is required to predict future observations. At test time, the model receives only the initial frame and action as input and autoregressively generates the subsequent video sequence. This setup allows us to test not only whether the model can predict short-term dynamics accurately, but also whether it can maintain coherent and stable rollouts over longer horizons through autoregressive generation. 

We evaluate the fidelity of learned dynamics on Meta-World, as summarized in Table \ref{tab:video_prediction}. 
In terms of model scale, World4RL has 330M parameters, which is comparable to NWM (320M)~\cite{bar2024navigationworldmodels} and remains smaller than iVideoGPT (430M)~\cite{wu2024ivideogpt}, while DiWA~\cite{chandra2025diwa} is considerably lighter at 40M parameters. 
Despite not being the largest model, World4RL consistently achieves the lowest FVD, FID, and LPIPS scores under both policy and random rollouts, significantly outperforming other methods. 
In particular, DiWA exhibits notably poor quantitative results on our multi-task dataset setting, and even its single-task variant (DiWA-ST) still lags far behind. 
These results suggest that the gains of World4RL stem not merely from model scale, but from the stronger temporal consistency and visual fidelity enabled by the design of architectures.

\begin{table}[t]
\centering
\caption{Quantitative results on video prediction. “ST” denotes single-task training and evaluation.}
\setlength{\tabcolsep}{3pt}       
\begin{tabular}{lcccccc}
\toprule
\multirow{2}{*}{Model} & \multicolumn{2}{c}{FVD $\downarrow$} & \multicolumn{2}{c}{FID $\downarrow$} & \multicolumn{2}{c}{LPIPS $\downarrow$} \\
\cmidrule(lr){2-3} \cmidrule(lr){4-5} \cmidrule(lr){6-7}
 & Policy & Random & Policy & Random & Policy & Random \\
\midrule
\textbf{\textsc{World4RL} (Ours)} & \textbf{326.5} & \textbf{400.1} & \textbf{17.1} & 23.4 & \textbf{0.0192} & \textbf{0.0246} \\
\textbf{NWM}~\cite{bar2024navigationworldmodels} & 547.4 & 851.9 & 30.5 & 34.9 & 0.0268 & 0.0259 \\
\textbf{iVideoGPT}~\cite{wu2024ivideogpt} & 450.3 & 531.3 & 18.7 & \textbf{20.7} & 0.0256 & 0.0283 \\
\textbf{DiWA}~\cite{chandra2025diwa} & 803.6 & 1231.0 & 62.9 & 96.5 & 0.0804 & 0.1364 \\
\textbf{DiWA (ST)} & 644.8 & 880.2 & 35.1 & 52.8 & 0.0523 & 0.0596 \\
\bottomrule
\end{tabular}
\label{tab:video_prediction}
\end{table}

\begin{figure}
    \centering
    \includegraphics[width=\linewidth]{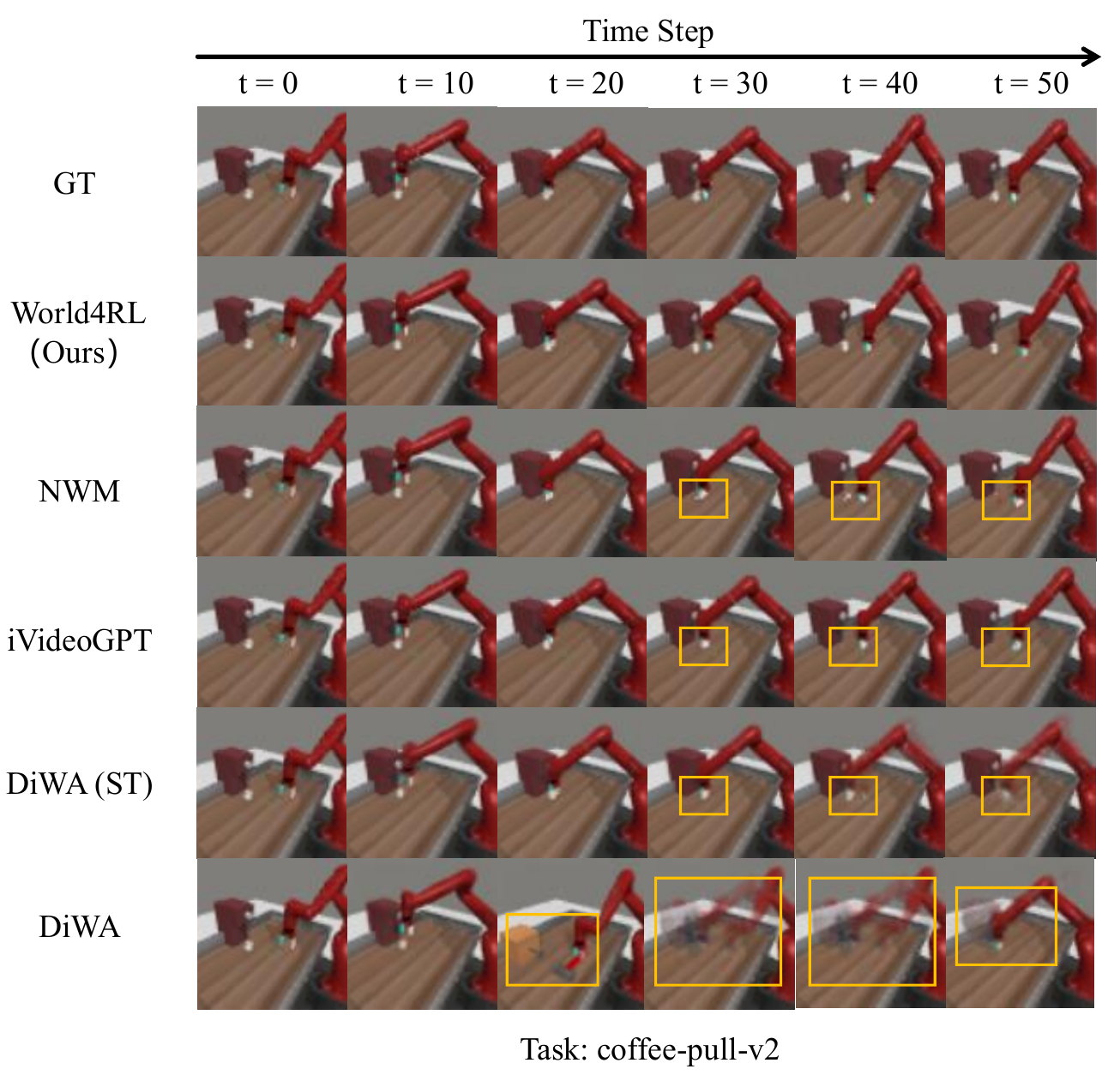}
    \caption{Visualization of predicted rollouts on the Coffee-Pull-v2 task. The ground truth (GT) trajectory corresponds to a failed execution, where the robot does not successfully pull the cup. World4RL accurately models this failure trajectory, faithfully capturing the underlying dynamics, while baseline models (NWM~\cite{bar2024navigationworldmodels}, iVideoGPT~\cite{wu2024ivideogpt}, and DiWA~\cite{chandra2025diwa}) incorrectly generate successful executions.}
    \captionsetup{skip=3pt}
    \label{fig:worldmodel}
\end{figure}

In addition to quantitative metrics, we provide visualizations in Fig.~\ref{fig:worldmodel}. World4RL produces more coherent and physically consistent rollouts, closely matching ground-truth trajectories, while baseline models often generate blurrier predictions or inconsistent dynamics. Notably, when given failed execution trajectories, World4RL can still faithfully model the underlying failure dynamics, whereas DiWA not only suffers from blurry and inconsistent rollouts but occasionally generates scenes from other tasks, underscoring its inability to generalize to multi-task settings. Together, these findings confirm that our diffusion world models provide superior capacity to capture fine-grained robotic interactions and long-horizon dynamics compared with other architectures.

\subsection{Can World4RL Enhance Policy Learning?}

We next investigate whether World4RL can facilitate policy learning and outperform existing approaches. Our baselines include imitation learning, offline RL, and world-model-based methods. To ensure a fair comparison, we adapt TD-MPC2 to the same offline world-model learning setting as DiWA and World4RL: it first learns a world model from the offline dataset and then performs policy optimization without any additional online environment interaction. In addition, all world-model-based methods are trained using the same data sources.

For a realistic evaluation, although Meta-World provides dense rewards, we instead use sparse success signals, which better reflect robotic settings where dense reward shaping is often unavailable. For imitation learning, the training set contains 50 expert demonstrations. Offline RL methods are trained on 50 expert trajectories plus 150 BC-policy rollouts. World-model-based methods use no additional online interaction and are optimized purely through simulator interaction, with the world model trained on 230 offline trajectories per task (50 expert demos, 150 BC-policy rollouts, and 30 random rollouts).

\begin{table*}[t]
\centering
\caption{Success rate of different methods on Meta-World benchmark over 3 seeds. The notation $\uparrow$ \textcolor{gray}{n} indicates the absolute improvement over the pre-trained policy.}
\captionsetup{skip=3pt}
\label{tab:rl_performance}
\setlength{\tabcolsep}{4pt}
\begin{tabular}{lcccccccc}
\toprule
\multirow{2}{*}{\textbf{Task}} 
& \multicolumn{2}{c}{\textbf{Imitation Learning}} 
& \multicolumn{2}{c}{\textbf{Offline Reinforcement Learning}} 
& \multicolumn{4}{c}{\textbf{World-Model-Based Methods}} \\
\cmidrule(lr){2-3} \cmidrule(lr){4-5} \cmidrule(lr){6-9}
& \textbf{BC (Base)} 
& \textbf{DP~\cite{ChiFDXCBS23}} 
& \textbf{TD3+BC~\cite{fujimoto2021minimalist}} 
& \textbf{IQL~\cite{DBLP:conf/iclr/KostrikovNL22}} 
& \textbf{IRASim~\cite{FangqiIRASim2024}} 
& \textbf{DiWA~\cite{chandra2025diwa}} 
& \textbf{TD-MPC2~\cite{hansen2024tdmpc2}} 
& \textbf{\textsc{World4RL} (Ours)} \\
\midrule
coffee-pull-v2   & 47 $\pm$ 7 & 34 $\pm$ 7 & 57 $\pm$ 13 & 47 $\pm$ 9  & 55 $\pm$ 6 & 49 $\pm$ 10 & 60 $\pm$ 7 & \textbf{68 $\pm$ 5} $\uparrow$ \textcolor{gray}{21} \\
soccer-v2        & 18 $\pm$ 2 & 19 $\pm$ 4 & 22 $\pm$ 8  & 14 $\pm$ 7  & 28 $\pm$ 5 & 21 $\pm$ 6 & 24 $\pm$ 5 & \textbf{31 $\pm$ 4} $\uparrow$ \textcolor{gray}{13} \\
hammer-v2        & 79 $\pm$ 5 & 15 $\pm$ 6 & 89 $\pm$ 4  & 73 $\pm$ 10 & 82 $\pm$ 5 & 83 $\pm$ 7 & 85 $\pm$ 5 & \textbf{91 $\pm$ 4} $\uparrow$ \textcolor{gray}{12} \\
door-lock-v2     & 74 $\pm$ 5 & 86 $\pm$ 8 & 82 $\pm$ 6  & 69 $\pm$ 11 & 78 $\pm$ 5 & 88 $\pm$ 7 & 85 $\pm$ 4 & \textbf{92 $\pm$ 5} $\uparrow$ \textcolor{gray}{14} \\
lever-pull-v2    & 31 $\pm$ 5 & 49 $\pm$ 5 & 39 $\pm$ 11 & 24 $\pm$ 3  & 33 $\pm$ 5 & 50 $\pm$ 9 & 39 $\pm$ 6 & \textbf{52 $\pm$ 8} $\uparrow$ \textcolor{gray}{21} \\
handle-pull-v2   & 60 $\pm$ 6 & 67 $\pm$ 11& 57 $\pm$ 6  & 25 $\pm$ 6  & 66 $\pm$ 6 & 68 $\pm$ 6 & 67 $\pm$ 5 & \textbf{71 $\pm$ 4} $\uparrow$ \textcolor{gray}{11} \\
\midrule
\textbf{Average SR} & 51.5 & 45.0 & 57.7 & 42.0 & 57.0 & 59.8 & 60.0 & \textbf{67.5} $\uparrow$ \textcolor{gray}{16} \\
\bottomrule
\end{tabular}
\end{table*}


Table~\ref{tab:rl_performance} reports the success rates on six Meta-World tasks. World4RL achieves the best average success rate of 67.5\%, outperforming all other baselines. Compared with imitation learning baselines, World4RL shows especially large gains on more challenging tasks such as \textit{coffee-pull-v2}, \textit{soccer-v2}, and \textit{lever-pull-v2}, suggesting that imitation alone is insufficient when demonstrations are limited. Compared with offline RL methods, World4RL also shows consistent advantages. Although TD3+BC and IQL can improve over pure imitation in some cases, their performance remains constrained by the fixed dataset. Among world-model-based baselines, World4RL also delivers the best overall performance. While IRASim is competitive on several tasks, it relies on computationally expensive test-time planning. DiWA and TD-MPC2 achieve stronger average performance than IRASim, but still lag behind World4RL. This result suggests that high-fidelity diffusion-based dynamics modeling is particularly beneficial for stable and effective policy optimization.

\begin{figure}[ht]
\centering
\includegraphics[width=0.95\linewidth]{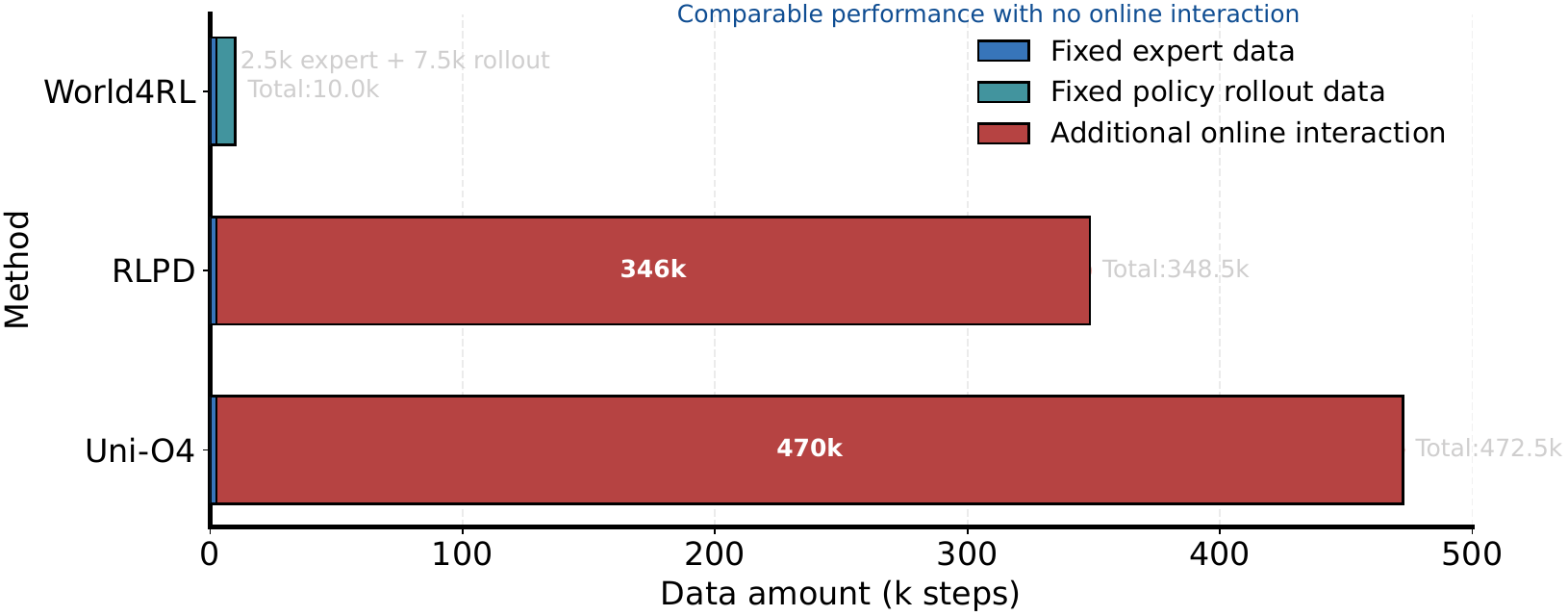}
\caption{Comparison of online sample efficiency. World4RL achieves comparable performance on fixed datasets, whereas RLPD and Uni-O4 require over additional 300k online steps.}
\label{fig:sample_efficiency}
\end{figure}

To further examine sample efficiency, we also compare World4RL with two representative offline-to-online approaches, Uni-O4\cite{DBLP:conf/iclr/LeiHLH0X24} and RLPD\cite{DBLP:conf/icml/BallSKL23}, both of which rely on substantial online interaction. In contrast, World4RL is trained entirely on fixed datasets without any online samples. As illustrated in Fig. \ref{fig:sample_efficiency}, World4RL already achieves comparable or superior performance with only expert and policy rollout data, while RLPD and Uni-O4 require 346k and 470k online steps, respectively, to reach the same level. This demonstrates the strong sample efficiency of World4RL, making it particularly suitable for real-robot deployment where online interaction is expensive and limited.


\subsection{Can World4RL Transfer to Real-World Robots well?}





\begin{figure}[ht]
    \centering

    \begin{subfigure}{0.49\linewidth}
        \includegraphics[width=\linewidth]{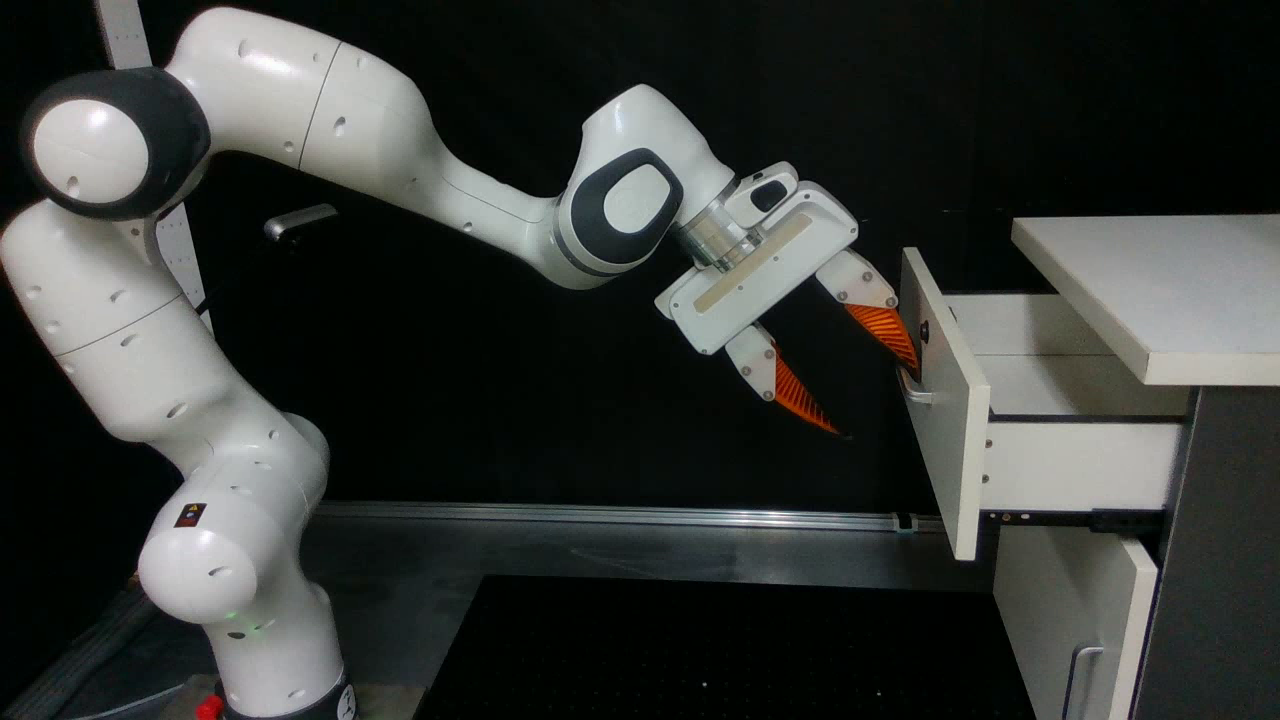}
        \caption*{Open Drawer} 
    \end{subfigure}\hfill
    \begin{subfigure}{0.49\linewidth}
        \includegraphics[width=\linewidth]{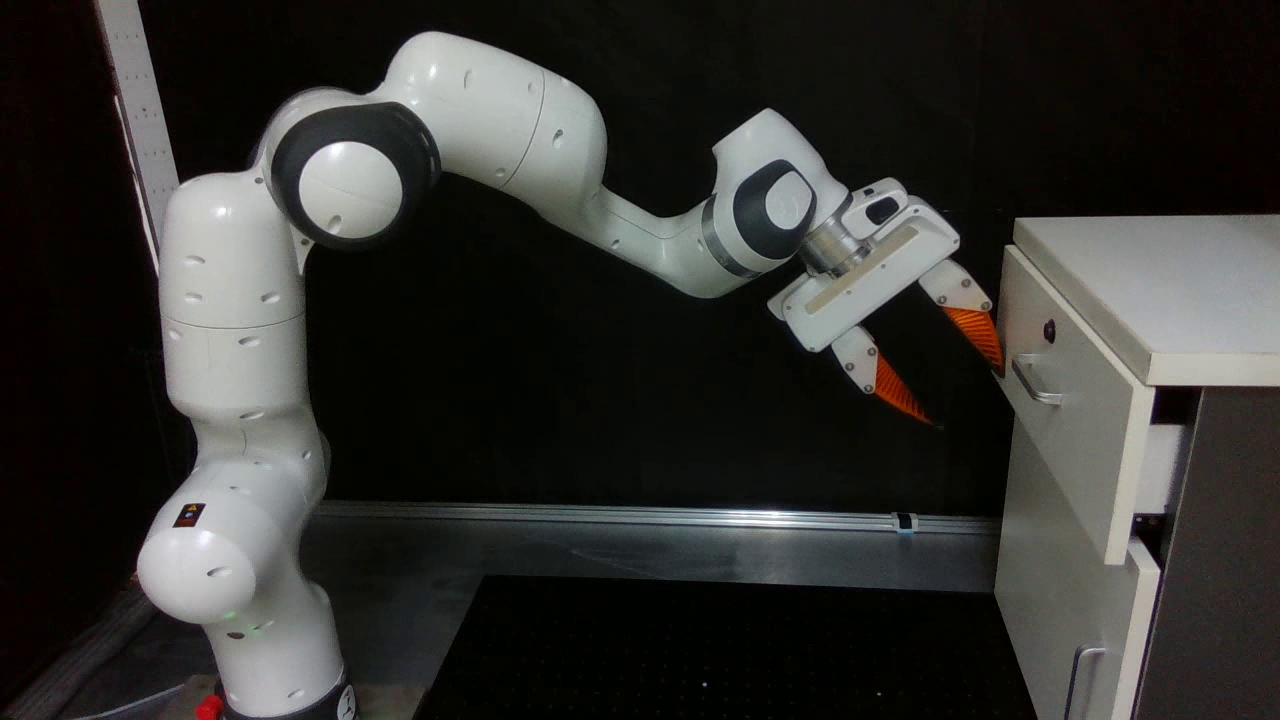}
        \caption*{Close Drawer}
    \end{subfigure}

    \begin{subfigure}{0.49\linewidth}
        \includegraphics[width=\linewidth]{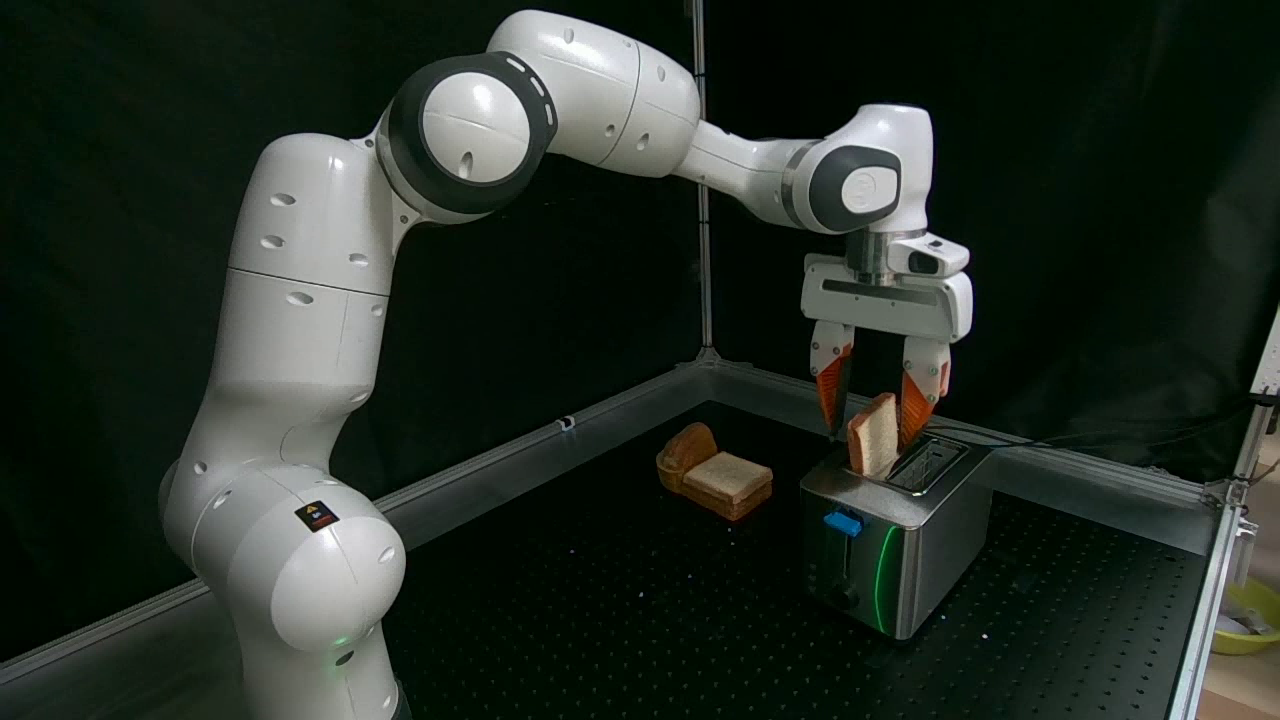}
        \caption*{Pick Bread In}
    \end{subfigure}\hfill
    \begin{subfigure}{0.49\linewidth}
        \includegraphics[width=\linewidth]{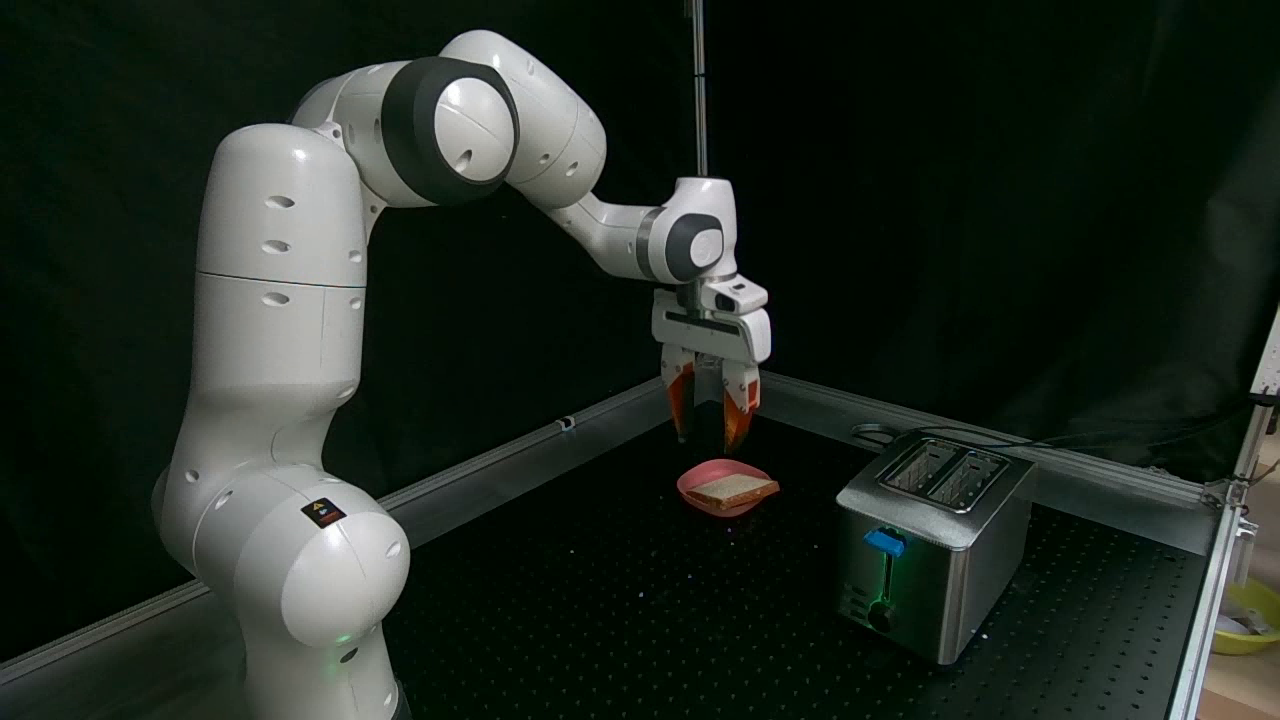}
        \caption*{Pick Bread Out}
    \end{subfigure}

    \begin{subfigure}{0.49\linewidth}
        \includegraphics[width=\linewidth]{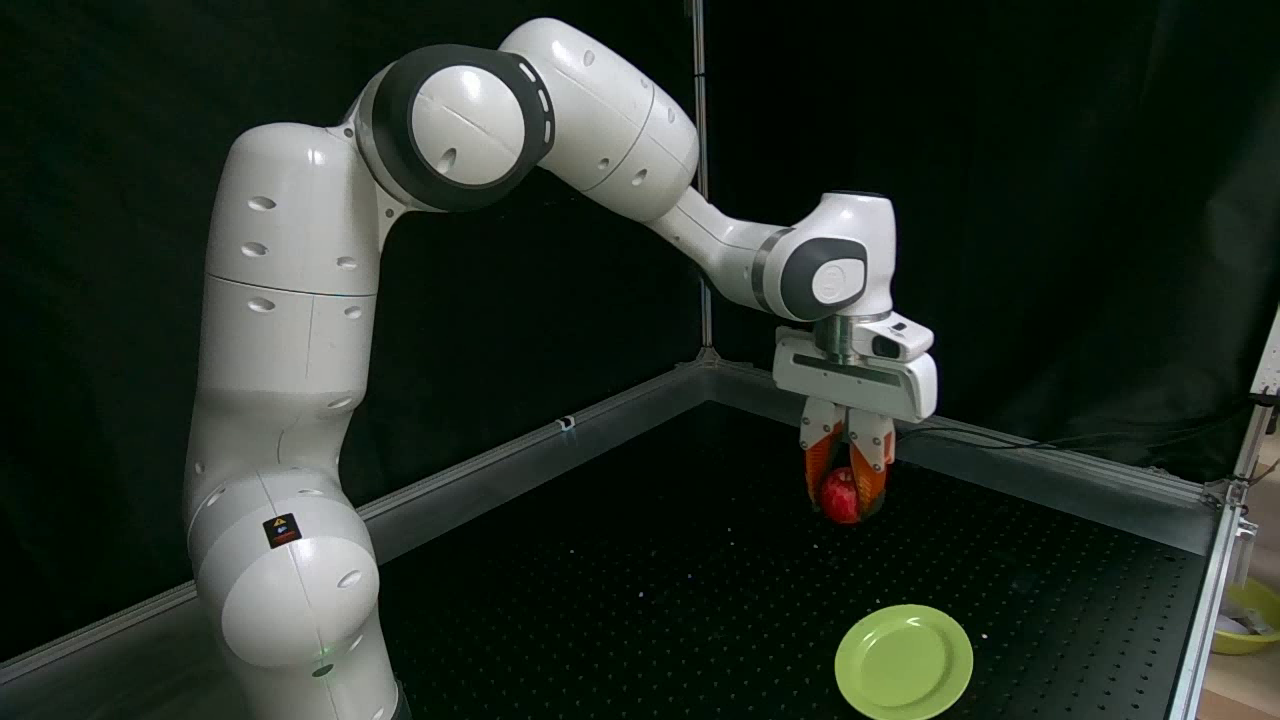}
        \caption*{Pick Apple}
    \end{subfigure}\hfill
    \begin{subfigure}{0.49\linewidth}
        \includegraphics[width=\linewidth]{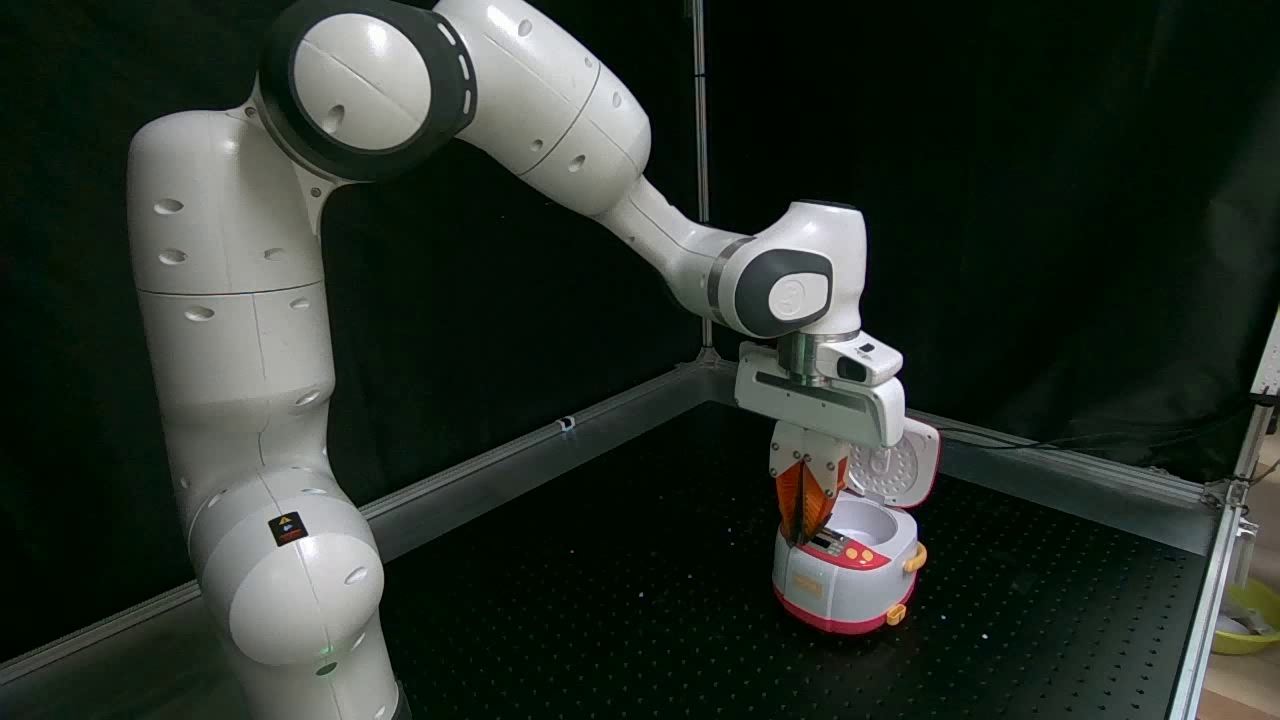}
        \caption*{Press Button}
    \end{subfigure}
    \caption{Real World Tasks}
    \captionsetup{skip=3pt}
    \label{fig:realtasks}
\end{figure}

We evaluate World4RL on six real-world manipulation tasks using a Franka Emika Panda robot, shown in Fig. \ref{fig:realtasks}. Following the HIL-SERL\cite{luo2025precise} protocol, we collected data via teleoperation with a space mouse. For each task, the training dataset comprised 50 human expert demonstrations, 50 trajectories generated by a pre-trained policy, and 50 trajectories from a random policy. Following the methodology in Sec.~\ref{sec:world4rl}, World4RL first pre-trains the policy and the diffusion world model, and then optimizes the policy with imagined rollouts, without requiring any additional real-world interaction. 

During evaluation, the initial scene configuration and robot starting pose are fixed for each task, and we execute 20 rollouts in the physical environment to measure success rates. We compare World4RL against the initial pre-trained policy and diffusion policy to assess whether our framework can deliver consistent improvements in real-world policy performance.  


\begin{table}[ht]
\centering
\caption{Real-world success rates across 6 manipulation tasks (20 trials per task).}
\captionsetup{skip=3pt}
\label{tab:real_robot_results}

\begin{tabular}{lccc}
\toprule
\textbf{Task} & \textbf{BC (Base)} & \textbf{DP}\cite{ChiFDXCBS23} & \textbf{\textsc{World4RL} (Ours)} \\
\midrule
Pick bread out & 13/20 & 19/20 & \textbf{20/20} \\
Pick apple     & 8/20  & 15/20 & \textbf{19/20} \\
Press button   & 12/20 & 16/20 & \textbf{18/20} \\
Put bread in   & 12/20 & \textbf{18/20} & 16/20 \\
Open drawer    & 12/20 & 18/20 & \textbf{19/20} \\
Close drawer   & 15/20 & \textbf{20/20} & \textbf{20/20} \\
\midrule
\textbf{Average SR} & 68.3\% & 88.3\% & \textbf{93.3\%} $\uparrow$ \textcolor{gray}{25} \\
\bottomrule
\end{tabular}
\end{table}

Table~\ref{tab:real_robot_results} reports the success rates across six real-world tasks. World4RL achieves the highest average success rate of 93.3\%, significantly outperforming imitation learning methods. Beyond achieving higher success rates, we also observe that policies fine-tuned with World4RL tend to execute tasks more decisively, completing pick-and-place behaviors quickly and accurately; for example, in the \textit{put bread in} task, the fine-tuned policy promptly performs the grasping and placing actions, whereas BC policy and diffusion policy\cite{ChiFDXCBS23} often show hesitation or linger in intermediate states without committing to task completion.

Our World4RL experiments are conducted on NVIDIA A800 GPUs (40GB). 
Pre-training the world model takes 20 hours on 4 A800 GPUs, while single-task policy refinement requires around 6 hours on a single A800 GPU. 

\section{Ablation Study}

We conduct ablation studies to analyze two key design choices in World4RL: the action encoding strategy and the training data composition of the world model.

\paragraph{Ablations on Action encoding}  
Action representation plays a critical role in world model learning, as it serves as the interface between the policy and the dynamics model. We compare our proposed two-hot encoding with several alternative strategies, including linear projection, one-hot discretization (\textbf{One-hot}~\cite{alonso2024diffusionworldmodelingvisual}, latent embeddings (\textbf{VQ-VAE}~\cite{lee2024behavior}), and tokenization (\textbf{FAST}~\cite{pertsch2025fast}). 

As shown in Table~\ref{tab:action_encoding}, two-hot encoding consistently achieves the best performance across FVD, FID, and LPIPS. Compared with other representations, two-hot encoding preserves action continuity while introducing a lightweight discrete structure, avoiding the reconstruction errors introduced by tokenization or latent compression. This enables more accurate dynamics modeling and more stable downstream reinforcement learning.

\begin{table}[htbp]
\centering
\caption{Comparison of different action encoding strategies on Meta-World video prediction.}
\label{tab:action_encoding}
\setlength{\tabcolsep}{3pt}       
\begin{tabular}{lcccccc}
\toprule
\multirow{2}{*}{Model} & \multicolumn{2}{c}{FVD $\downarrow$} & \multicolumn{2}{c}{FID $\downarrow$} & \multicolumn{2}{c}{LPIPS $\downarrow$} \\
\cmidrule(lr){2-3} \cmidrule(lr){4-5} \cmidrule(lr){6-7}
 & Policy & Random & Policy & Random & Policy & Random \\
\midrule
\textbf{Two-hot (Ours)}   & \textbf{326.5} & \textbf{400.1} & \textbf{17.07} & \textbf{23.43} & \textbf{0.0192} & \textbf{0.0246} \\
\textbf{One-hot}\cite{alonso2024diffusionworldmodelingvisual}   & 350.3 & 471.5 & 18.52 & 26.24 & 0.0193 & 0.0257 \\
\textbf{Linear}\cite{bar2024navigationworldmodels} & 353.4 & 514.0 & 17.85 & 23.83 & 0.0218 & 0.0250 \\
\textbf{FAST}\cite{pertsch2025fast}     & 407.0 & 748.0 & 28.92 & 36.52 & 0.0284 & 0.0409 \\
\textbf{VQ-VAE}\cite{lee2024behavior}    & 525.6 & 860.0 & 28.60 & 43.25 & 0.0506 & 0.0633 \\
\bottomrule
\end{tabular}
\end{table}




\paragraph{Ablations on policy optimization performance}

We study two practical design choices that are important for stable and effective policy optimization in the world model: controlled policy exploration and broader action-space coverage during world model training. Specifically, we compare our full method against two ablations: \textbf{w/o action std clipping}, which removes the variance constraint on the Gaussian policy during PPO training, and \textbf{w/o random rollouts}, which excludes uniformly sampled random transitions from world model training.

\begin{figure}[ht]
\centering
\includegraphics[width=0.95\linewidth]{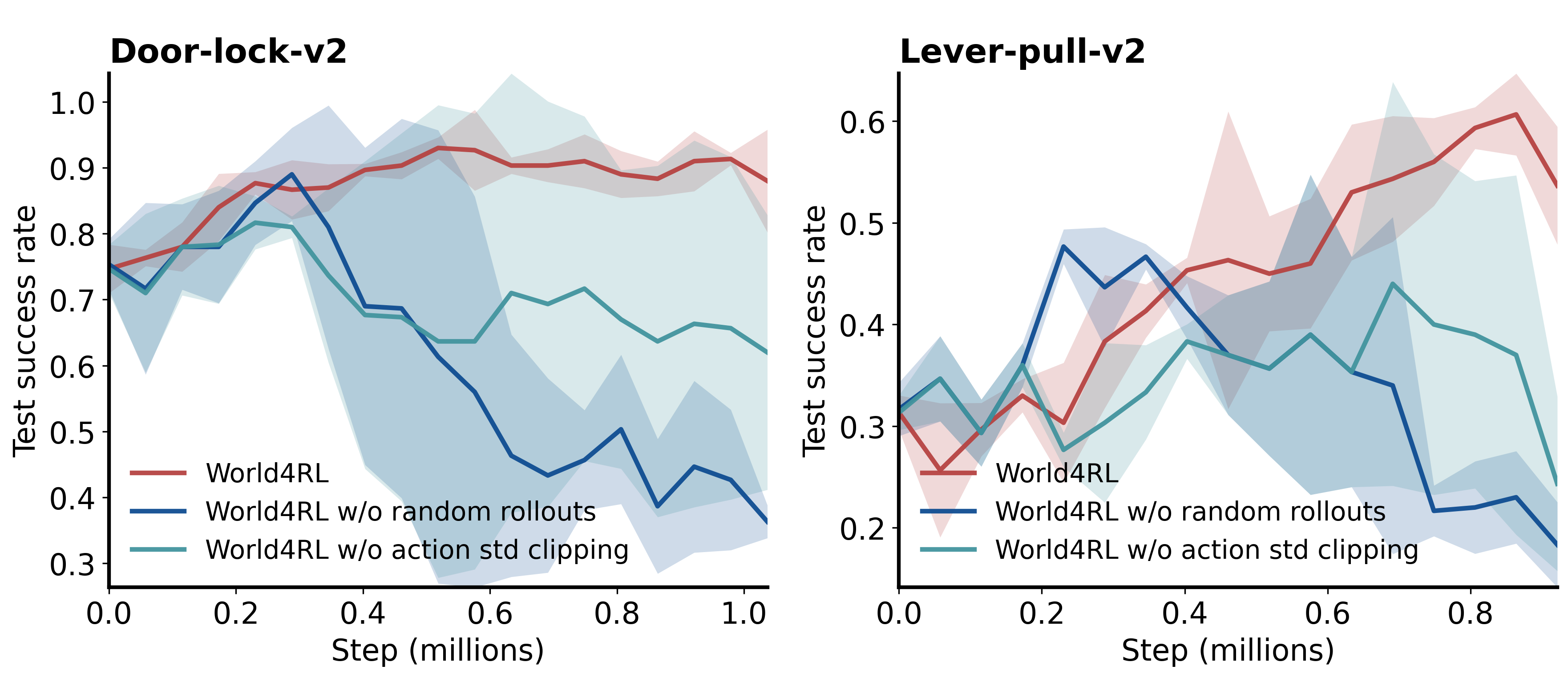}
\caption{Ablations on policy optimization performance. We compare the full method with two variants: \textbf{w/o action std clipping} and \textbf{w/o random rollouts}.}
\captionsetup{skip=3pt}
\label{fig:policy_opt_ablation}
\end{figure}

Figure~\ref{fig:policy_opt_ablation} shows that removing either component degrades policy performance during RL training. In particular, both \textbf{w/o action std clipping} and \textbf{w/o random rollouts} lead to clear performance drops on \textit{Door-lock-v2} and \textit{Lever-pull-v2}. Moreover, \textbf{w/o action std clipping} exhibits notably larger variance throughout training, indicating that controlled policy exploration is important not only for final performance but also for optimization stability. These results show that controlled exploration during policy optimization and broader action-space coverage during world model training are both critical for reliable imagined rollouts and strong final policy performance.

\section{Conclusion and Future Work}
In this work, we proposed World4RL, a framework that systematically incorporates diffusion models into reinforcement learning for robotic manipulation. Experimental results demonstrate that World4RL not only serves as a high-fidelity world model capable of accurately modeling trajectories, but also functions as a real-time simulator that that enables efficient policy refinement under sparse reward conditions. These findings highlight the potential of diffusion models as a unifying bridge between visual prediction and reinforcement learning, facilitating consistent improvement of pre-trained policies beyond imitation learning.


At the same time, we emphasize that this work mainly establishes the feasibility of this direction rather than exhausting its full potential. Due to computational constraints, the current implementation adopts relatively moderate visual resolution and model capacity, which may limit the fidelity of imagined rollouts. We expect that improving visual detail and generative expressiveness will further enhance the quality of world modeling and lead to stronger downstream policy performance.
In the future, we plan to explore two main directions: enhancing the fidelity of world models through richer visual representations, and developing more robust reinforcement learning algorithms that can effectively learn under imperfect learned dynamics.









\bibliographystyle{IEEEtran}
\bibliography{ref}

\end{document}